\documentclass{article}

\usepackage[preprint]{neurips_data_2022}

\usepackage[utf8]{inputenc} 
\usepackage[T1]{fontenc}    
\usepackage{hyperref}       
\usepackage{xurl}            
\usepackage{booktabs}       
\usepackage{amsfonts}       
\usepackage{nicefrac}       
\usepackage{microtype}      
\usepackage{xcolor}         
\usepackage[pdftex]{graphicx}
\usepackage{multirow}
\usepackage{array}
\usepackage{float}

\newcommand{\eg}{{\textit{e.g.}}}

\title{A Large-Scale Annotated Multivariate Time Series Aviation Maintenance Dataset from the NGAFID}

\author{%
  Hong Yang \\
  Rochester Institute of Technology\\
  Rochester, NY 14623 \\
  \texttt{hy3134@rit.edu} \\
   \And
   Travis Desell\\
   Rochester Institute of Technology \\
   Rochester, NY 14623\\
   \texttt{tjdvse@rit.edu} \\
}

\begin{document}

\maketitle

\begin{abstract}

This paper presents the largest publicly available, non-simulated, fleet-wide aircraft flight recording and maintenance log data for use in predicting part failure and maintenance need. We present 31,177 hours of flight data across 28,935 flights, which occur relative to 2,111 unplanned maintenance events clustered into 36 types of maintenance issues. Flights are annotated as before or after maintenance, with some flights occurring on the day of maintenance.  Collecting data to evaluate predictive maintenance systems is challenging because it is difficult, dangerous, and unethical to generate data from compromised aircraft. To overcome this, we use the National General Aviation Flight Information Database (NGAFID), which contains flights recorded during regular operation of aircraft, and maintenance logs to construct a part failure dataset. We use a novel framing of Remaining Useful Life (RUL) prediction and consider the probability that the RUL of a part is greater than 2 days. Unlike previous datasets generated with simulations or in laboratory settings, the NGAFID Aviation Maintenance Dataset contains real flight records and maintenance logs from different seasons, weather conditions, pilots, and flight patterns. Additionally, we provide Python code to easily download the dataset and a Colab environment to reproduce our benchmarks on three different models. Our dataset presents a difficult challenge for machine learning researchers and a valuable opportunity to test and develop prognostic health management methods. 

\end{abstract}

\section{Introduction}

Maintenance related issues are a notable risk in the operation single-engine fixed-wing aircraft \cite{goldman2002general} for private use, known as General Aviation (GA). With improvements in data collection and processing capabilities, there is now an opportunity to indirectly monitor the condition of a variety of components in GA aircraft for the purposes of Predictive Maintenance (PM).  

The domain of prognostics and health management (PHM) has shifted towards the use of more data driven approaches \cite{tsui2015prognostics}, sometimes also involving deep learning methods. These methods aim to predict the RUL, detect faults, or monitor the condition of a system and its parts. However, it is particularly difficult to collect data for these purposes. \cite{rezaeianjouybari2020deep} states that the data collection process is often too time consuming and too perfect (the data collected in laboratory conditions may not translate well to the real world). It is clear that there is a lack of high quality, real world data, yet there is a great need for such data to evaluate new PHM methodologies. 

Existing datasets for PHM often use simulated data \cite{liu2012user} or model systems in controlled environments, such as a power plant data in the 2015 Prognostics and Health Management Society Data Challenge. However, PHM methodologies can be extended to real world systems in largely uncontrolled environments, such as cars or aircraft. As vehicles collect more data, see \cite{arena2021predictive}, there is an opportunity to combine said data with unplanned maintenance records to create a novel RUL estimation dataset for predictive maintenance.

GA flight data is particularly challenging for PHM due to the nature of the aircraft sensors. Unlike commercial aircraft, GA aircraft only possess basic sensors for monitoring critical systems, such as the engine and battery, and flight instruments, for measuring air speed and altitude. A predictive model relying on such basic sensors for predicting RUL of specific parts must be both highly sensitive to the most subtle of changes and highly robust to noise. This is because a significant portion of the variance in flight data is explained by pilot action and not the condition of the airplane's parts. This poses a difficult problem for conventional approaches, such as auto encoders.

In this paper we present the NGAFID Aviation Maintenance Dataset for use in binary RUL estimation. This dataset is constructed using flight records and unplanned maintenance logs of a single flight school over multiple years. For each unplanned breakdown of a part, which lead to unplanned maintenance, there are multiple flights days before and after the maintenance event. From here, we create two groups of flight data, one with parts that are going to break within $x$ days and ones with new parts, which are not expected to break within $x$ days. This challenging PHM problem uses one of the largest real-world PHM datasets and it also poses a particularly difficult machine learning problem for time series classification and time series anomaly detection.

\subsection{Our Contributions}

\textbf{Data: \url{https://doi.org/10.5281/zenodo.6624956} and \url{https://www.kaggle.com/datasets/hooong/aviation-maintenance-dataset-from-the-ngafid}} A dask dataframe containing 31,177 hours of flight data across 28,935 flights, with a header csv that describes each flight and the associated maintenance file. The data was collected automatically by a flight school after each flight and uploaded to the NGAFID database. Flights occurred in a variety of seasons and weather conditions. All flights were flown with Cessna 172 aircraft. 23 sensors record data every second, resulting in more than 100 million rows of flight data with a total size of 4.3 GB.

\textbf{Benchmarks: \url{https://github.com/hyang0129/NGAFIDDATASET}} A repository containing helper code to automatically download and process the dataset and 2 Colab notebooks for replicating the benchmark experiments. Anyone can run the benchmark experiments with one click using the Colab notebooks in a web browser, with a free Linux environment including GPUs and TPUs provided by Google. This demonstrates how to use the dataset and replicate experiments, regardless of the replicator's hardware and software limitations. 

The above files are licensed under GNU General Public License V3.0.

\section{Related Work} 

\subsection{Aircraft and Aircraft Maintenance Datasets}

Publicly available aircraft maintenance datasets often use simulated data rather than data collected from real life events. Commercial Modular Aero-Propulsion System Simulation (C-MAPSS) \cite{liu2012user} is a commonly used simulation program for aircraft engines. These simulations are sometimes supplemented by real flight conditions, 
\eg ~\cite{arias2021aircraft}, where flight data is used as the input for the simulation. However, these approaches rely on synthetic data, which may not reflect the noisy nature of data collection. There are papers that reference real data, but they do not publicly release their datasets, \eg ~\cite{celikmih2020failure} and \cite{dangut2022rare}. The task of detecting anomalies from flight data was investigated by \cite{chu2010detecting}, however the dataset was generated from a flight simulator. \cite{yildirim2018aircraft} collects real world data from aircraft, but only during normal operation of the aircraft and not when there is a fault. A common difficulty in collecting real world failure data is the associated cost and safety risk; it would be unethical to ask pilots to fly faulty aircraft. \cite{yang2021predictive} presented a smaller version of this dataset for aviation maintenance. 

Aircraft flight data has also been used for other tasks, aside from maintenance prediction. \cite{klein1989estimation} used real data to estimate aerodynamic properties of aircraft. \cite{khadilkar2012estimation} used aircraft flight recordings to estimate fuel usage during the taxing stage. Aircraft flight data can also be used to improve estimation of the remaining range for the aircraft, \cite{randle2011improved}. In most PHM dataests, \cite{liu2012user}, the goal is to predict an estimate or probability density function of the RUL, prior to a fault occuring. Prior to this work, there were no publicly available fleet-wide maintenance records of this size and scope coupled with flight data recordings in both fault free and faulty states. 

\subsection{Time Series Datasets}

There are many time series classification datasets, such as \cite{bagnall2018uea}. These datasets are used to evaluate both deep learning methods, such as Inception Time, \cite{fawaz2020inceptiontime}, and non deep methods, such as Mini Rocket, \cite{dempster2021minirocket} and HIVE-COTE, \cite{lines2018time}. Time series classification could also include audio datasets, such as \cite{6637622}, but such datasets belong in their own domain of audio processing. As shown by \cite{bagnall2018uea}, a collection of time series datasets will contain very diverse types of data, such as motion sensor data, electroencephalogram data, traffic data and more. Few time series datasets exist that contain as many training examples with so many data points per training example. 






\subsection{Classification Methods}

Several methods have been developed for MTS classification, for a review see \cite{fawaz2019deep}. Notable non-deep learning methods include distance based k-nearest neighbors by \cite{orsenigo2010combining} and Dynamic Time Warping KNN by \cite{seto2015multivariate}. For deep learning methods, well performing MTS classifiers tend to utilize some combination of Recurrent Neural Network (RNN) and Convolutional Neural Network (CNN) methods, \cite{karim2017lstm}, or CNN only methods  \cite{wang2017time}. However, RNN methods struggle with long sequences due to the vanishing gradient problem, as mentioned by \cite{le2016quantifying}. CNN only methods can provide strong results for MTS classification, as shown by \cite{assaf2019mtex}, but may struggle when relevant features are temporally sparse and related.  \cite{yang2021predictive} also used multi-headed self-attention methods. 

\section{Dataset Description}

\subsection{Data Acquisition}

The full dataset is constructed using flight recordings from the NGAFID database and maintenance records from MaintNet. Each maintenance record describes the type of maintenance performed on an aircraft and the time period when maintenance occurred. From this information, we extract that aircraft's flights occurring before and after the maintenance period. Only the first 5 flights before and after and any flights during maintenance are extracted. All maintenance activities were unplanned and done on request. However, when issues are detected, FAA policy forbids the plane from flying. This means that issues are always detected sometime during or after the first flight before maintenance. Each flight record contains readings from 23 different sensors every second. A description of the sensors can be found in Table \ref{Sensor}. All flight data recordings come from the same aircraft model, the Cessna 172.

\begin{table}[t!]
\centering
\small
{\setlength{\extrarowheight}{2pt}%
\begin{tabular}{|l|l|}
\hline

  Column Name & Description  \\ \hline

volt1 & Main electrical system bus voltage (alternators and main battery) \\ \hline 
volt2 & Essential bus (standby battery) bus voltage \\ \hline 
amp1 & Ammeter on the main battery (+ charging, - discharging)  \\ \hline
amp2 & Ammeter on the standby battery (+ charging, - discharging)  \\ \hline
FQtyL & Fuel quantity left  \\ \hline
FQtyR & Fuel quantity right \\ \hline
E1 FFlow & Engine fuel flow rate \\ \hline
E1 OilT & Engine oil temperature \\ \hline
E1 OilP & Engine oil pressure \\ \hline
E1 RPM & Engine rotations per minute \\ \hline
E1 CHT1 & 1st cylinder head temperature \\ \hline
E1 CHT2 & 2nd  cylinder head temperature \\ \hline
E1 CHT3 & 3rd  cylinder head temperature \\ \hline
E1 CHT4 & 4th  cylinder head temperature \\ \hline
E1 EGT1 & 1st Exhaust gas temperature  \\ \hline 
E1 EGT2 & 2nd Exhaust gas temperature \\ \hline 
E1 EGT3 & 3rd Exhaust gas temperature \\ \hline 
E1 EGT4 & 4th Exhaust gas temperature  \\ \hline 
OAT & Outside air temperature \\ \hline 
IAS & Indicated air speed \\ \hline 
VSpd & Vertical speed\\ \hline 
NormAc & Normal acceleration \\ \hline 
AltMSL & Altitude miles above sea level \\ \hline

\end{tabular}}

\caption{Description of the data collected by aircraft sensors}
\label{Sensor}

\end{table}

The NGAFID serves as a repository for general aviation flight data, with a web portal for viewing and tracking flight safety events for individual pilots as well as for fleets of aircraft~\cite{karboviak2018classifying}. The NGAFID currently contains over 900,000 hours of flight data generated by over 780,000 flights by 12 different types of aircraft, provided by 65 fleets and individual users, resulting in over 3.15 billion per second flight data records across 103 potential flight data recorder parameters. Five years of textual maintenance records from a fleet which provided data to the NGAFID have been clustered by maintenance issue type and then validated by domain experts for the MaintNet project, see~\cite{akhbardeh-etal-2020-maintnet}. 

MaintNet's maintenance record logbook data was clustered into 36 different maintenance issue types. The count of flights per issue type is located in table \ref{classes} in Appendix A. Although some maintenance issues occur very rarely, all maintenance issues with flight data are included in the full dataset. It is important to note that the NGAFID collects real flight data from aircraft flying with potentially faulty parts (as is the case for any real world fleet of aircraft). This is because individual components may fail without causing catastrophic failure of the aircraft during regular operation. The collection of this data poses no additional safety risk to the pilots because data collection occurs for all flights performed. 

\subsection{Dataset Subset for Benchmarking}

Attempting to train a model on the full dataset is particularly difficult. This is because of two main issues. The first is the under representation of certain classes. For example, the spark plug related issue contains only 15 flights. These classes are included in the dataset to allow for future researchers to train models that address the class imbalance issue, as class imbalance is an important area of machine learning research \cite{johnson2019survey}. The second is that each flight contains a significant amount of data, which makes it hard to regularize a model trained on the data. In the spark plug example, each of the 15 flights contain more than 40,000 data points, which would be used to predict a single class out of 36. Regularization is another important research area of machine learning, \cite{kukavcka2017regularization}, but it is beyond the scope of this paper. These two challenges are inevitable when collecting real world data, as it is impossible to guarantee that each part in a real world aircraft fails at the same rate. 

A subset of the full dataset is used in this paper to benchmark time series classification methods and provide a baseline. Since addressing class imbalance and regularization are beyond the scope of this paper, the subset is designed to minimize the impact of those two issues. By limiting the subset to only classes that have at least 50 flights, it removes flights where generalization is a problem and reduces the impact of class imbalance. The subset also defines the binary RUL problem as $P(RUL > x)$ with $x$ as 2 days. We label flights within 2 days before maintenance as $P(RUL > x) = 0$ and flights after 2 days after maintenance as $P(RUL > x) = 1$, as such flights contain brand new parts. 

The data subset is defined as all flights within two days of the maintenance period, excluding any flights during the maintenance period. Furthermore, any flights belonging to classes that would have fewer than 50 flights before or after the maintenance period are also excluded. This leaves us with a total of 5844 flights after maintenance and 5602 flights before maintenance in 19 different classes of maintenance issues. The count of flights per issue type is located in table \ref{classes} in Appendix A.

\subsection{Data Preprocessing}

The full dataset is provided without any preprocessing and contains the full flight data for each flight. Some sensors recorded NaN values at certain time steps. This can be caused by many factors, including but not limited to initialization of the sensor at the beginning of the flight, failure of the sensor during flight, and failure of the recording system during flight. Approximately 1\% of all datapoints contain NaN values. We decided to present future researchers with NaN values as replacing such values with another, such as 0, would change the meaning. 

For training a model in our benchmark experiments, the data subset is scaled. All values are scaled via MinMax, with a scaled minimum value of 0 and a scaled maximum value of 1. Minimums and maximums for each channel of data were calculated using all of the data. Please note that the full dataset is not distributed with its values scaled, which allows for examination of different normalization techniques by users of the dataset.

\subsection{Visualization}

\begin{figure}[t]
    \centering
    \includegraphics[width=14cm]{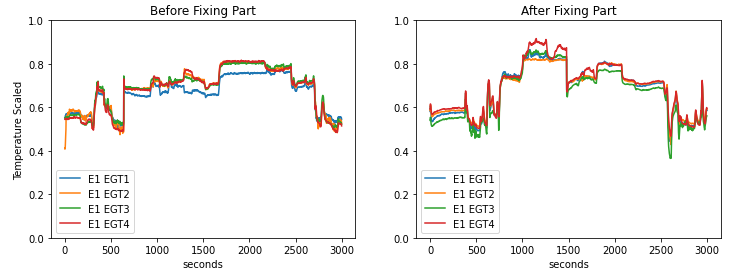}
    \caption{Exhaust Gas Temperature before and after Maintenance for Intake Gasket Issue. This shows that the sensor readings before an issue occurs and after it is fixed is largely the same. Notice that the EGT values are mostly in line with each other, but diverge at certain times. It is not obvious where an anomaly occurs, nor is it obvious whether or not there is an issue. The difference in temperature values indicate a difference in flight plan as well. }
    \label{fig:example}
\end{figure}

Readings from 4 sensors from two flights are provided in figure \ref{fig:example}. Only the output from 4 sensors are included to keep the graph readable. Even with only 4 sensors, it should be clear that determining the probability of part failure for a human would be extremely difficult. Even with domain knowledge, it would only be possible to detect that an issue has occurred, but not that an issue is going to occur. 

\subsection{Data Privacy}

To ensure privacy, information regarding the serial number of the aircraft of the flight and the date of the flight have been removed from the publicly released dataset. Latitude and longitude are also removed from the flight. The specific text of the maintenance logs are also withheld. 

\subsection{Labels}

Labels are assigned to flights based on the date of the flight and the date of the maintenance. As per the Federal Aviation Administration regulations, aircraft are only permitted to operate in a safe manner. This means that maintenance issues and part failures that occurred prior to the maintenance date will be fixed after the maintenance date.

Part failures in the aircraft can be described as acute or chronic. Acute part failure  describes a complete failure of a part due to an unexpected event.  For example, if the aircraft collides with a small object, such as a bird, this would create data representing an acute failure and an instantaneous change in the remaining useful life. Chronic part failure describes gradual wear and tear that renders a part unsafe for flight, before scheduled maintenance can replace the part. For example, an intake gasket may wear down more quickly than expected, leading to a leakage that negatively impacts the safety of the aircraft.

It should be noted that dramatic decreases in remaining useful life are rare. Based on analysis of the intake gasket leak/damage class, there were 9085 cases were described as leaking and only 15 cases were described as torn. This may negatively impact anomaly detection methods and will be discussed in a later section. 

Because all maintenance records are related to unplanned maintenance, it is clear that the related parts reached the end of their remaining useful life prematurely. These unplanned maintenance events occurred outside of scheduled maintenance and can be considered anomalies in the expect wear and tear of the associated parts. 

\subsection{Problem Structure}

\subsubsection{Detection of Maintenance Issues}

The detection of maintenance issues and their associated part failure is quite challenging. We define this as $ 
\min_{\forall i \in I} P(RUL_i > x)$, where $I$ represents the set of all parts and $RUL_i$ is the remaining useful life of part $i$.

While this problem may resemble anomaly detection, it is important to note that a significant portion of variance in the data is caused by outside factors. Pilot actions can dramatically alter readings for every sensor on the aircraft, causing no two flights to be the same. Weather conditions, which may vary with altitude, are also inconsistent. For these reasons, an anomaly detection approach may find a large number of anomalous sections on any given flight. These anomalies may be caused by a maintenance issue or irregular flight activities. 

It should be clear that the data collected by aircraft sensors pose a greater challenge than sensors on laboratory or industrial assets. In those situations, the environment is much more controlled, as opposed to the environment of an aircraft in flight. They may also utilize dedicated sensors designed to collect data for the purpose of detecting specific issue. This is not the case for our dataset, as the sensors collect general flight data, which were not originally designed to collect data for maintenance issue prediction and binary RUL estimation.  

\subsubsection{Classification of Maintenance Issues}

An extension of the previous problem is to also classify the type of maintenance issue based on the maintenance record left by the mechanic. This problem is significantly more difficult for two reasons. First is the class imbalance, with some classes having thousands of flights and others having less than one hundred. Second is the similarity in how maintenance issues affect flight characteristics. This is because two different issues may cause very similar changes in flight characteristics.

Formally, this is defined as $\forall i \in I [  P(RUL_i > x) ]$. This is significantly more difficult than detection of the maintenance issue.

\subsubsection{Hierarchical Classification}

In addition to the class labels for specific maintenance issues, this dataset is provided with a simple hierarchy for maintenance issues. There are 5 groupings of maintenance issues: engine, baffle, oil, cylinder, and other. \cite{argyriou2006multi} has shown that hierarchies can provide benefits in terms of regularization, as they improve the information provided by the labels. These hierarchies are included for potential future research.   

\section{Benchmark Experiments} 

\subsection{Dataset and Task Definition}

Three different tasks are defined and all three tasks use the same subset of data. The first task is maintenance issue detection. Here, we assign after maintenance flights with a negative label and before maintenance flights with a positive label. The second task is maintenance issue classification, where we assign after maintenance flights with class 0 and before maintenance flights with their maintenance issue class (from 1 to 19). The third is combined detection and classification, where a network is trained on both tasks simultaneously. 

\subsection{Models}

\subsubsection{Mini Rocket}

Mini Rocket \cite{dempster2021minirocket} is an improvement over the original Rocket classifier, \cite{dempster2020rocket}. While the original Rocket transforms the input time series using a large number of random convolutional kernels and trains a linear classifier on the transformed features, Mini Rocket improves upon this by using a smaller, fixed set of kernels. This method has performed quite well on many time series datasets and trains much more quickly than deep learning methods. We evaluate a GPU implementation of Mini Rocket by \cite{tsai}.

\subsubsection{HIVECOTEv2}

We attempted to test HIVECOTEv2 \cite{middlehurst2021hive} on the NGAFID maintenance dataset using the sktime implementation by \cite{markus_loning_2022_6547489}. However, the training time for the model exceeded the maximum time allocated for Google Colab instances (24hrs). \cite{shifaz2020ts} notes that HIVECOTE scales polynomially with data, which could explain why it could not train in time on the NGAFID maintenance dataset. Not only is the NGAFID maintenance dataset large, it also has many time steps and many channels. No results for HIVECOTEv2 are included.  

\subsubsection{Inception Time}

Convolutional networks are very popular in the domain of computer vision, but can also perform well in time series classification. \cite{fawaz2020inceptiontime} proposed the InceptionTime model as an ensemble of five Inception models. Each Inception model is composed of two residual blocks, each containing three Inception modules, followed at the very end by a global average pooling layer and a dense classification head layer. Inception modules contain convolutions of various kernel sizes and a bottleneck layer. The residual blocks help mitigate the vanishing gradient issue by allowing for direct gradient flow \cite{he2016identity}. \cite{fawaz2020inceptiontime} noted that ensembling was necessary due to the high standard deviation in accuracy of single Inception models and the small size of time series datasets.

For this study, we evaluate the Inception model without ensembling, so references to InceptionTime refer to just the Inception model, without ensembling. This is because the compute cost is reduced five-fold allowing for a more efficient use of limited resources. 

\subsubsection{ConvMHSA}

Multi-Headed Self Attention (MHSA) modules were popularized by \cite{devlin2018bert} for usage in Natural Language Processing and by \cite{dosovitskiy2020image} for Computer Vision. The ConvMHSA model used for benchmarking is the same as the one in \cite{yang2021predictive}. The model implements attention layers that mimic the functionality of the encoder layers present in BERT \cite{devlin2018bert}. Instead of token embeddings, the model generates sequence embeddings with the use of 1D convolutions along the temporal dimension. These learnable sequence embeddings capture local relationships and to compress the MTS to a shorter length. It uses a series of 1D convolutions to reduce the temporal resolution from 4096 to 512 and then employs 4 stacked  MHSA encoder layers with 8 heads each and 64 dense units per head. The output is globally average pooled and fed to a dense layer for classification.


\subsubsection{Model Configuration}

All results reported were generated using a Google Colab instance with a v2-8 TPU. All models were trained for 200 epochs with 200 steps per epoch using a batch size of 128, with 5-fold cross validation. Flights are truncated to the last 4096 time steps and padded to be of the same size. Models used an Adam optimizer with a learning rate of 3e-5 for CONVMHSA and 1e-4 for InceptionTime. 

Mini-rocket was trained using a Google Colab instance with a Nvidia V100 GPU. Models were trained for 200 epochs with 143 steps and a batch size of 64, with 5 fold cross validation.  Flights are truncated to the last 4096 time steps and padded to be of the same size. Models used an Adam optimizer with a learning rate of 2.5e-5. 

Notebooks used to train the models are available on Github and can be run in Colab through a web browser. Colab notebooks can be exported as regular Jupyter notebooks to run the experiments locally on a GPU. See \url{https://github.com/hyang0129/NGAFIDDATASET} for benchmark instructions.

\subsection{Results}

A summary of results can be found in Table \ref{res}. Overall, models tend to overfit the data, but especially so on classifying the specific maintenance issue. This is because certain classes contain only 75 examples. MiniRocket performed relatively poorly compared to the deep learning models. If we compare the training loss and the validation loss, it appears that MiniRocket may have difficulty in generalization in the multi class case. It also seems to suffer from under fitting in the binary case. These results suggest that deep learning methods may have an advantage for this type of problem.

\begin{table}[t!]
\centering

{\setlength{\extrarowheight}{2pt}%
\begin{tabular}{|l|l|l|l|l|l|}

\hline

  Model & Task & Binary Acc. & Multiclass Acc. & Loss & Train Loss \\ \hline
  \multirow{3}{*}{ConvMHSA} & Binary & 76.0\% & N/A & 0.526 & 0.003 \\ \cline{2-6}
  & Multi & N/A & 52.8\%   & 2.168 & 1.097 \\ \cline{2-6}
  & Both & \textbf{76.1}\% & \textbf{56.3}\%  & 1.756 & 1.377 \\ \hline
    \multirow{3}{*}{InceptionTime} & Binary & 75.5\% & N/A & 0.569 & 0.214 \\ \cline{2-6}
  & Multi & N/A & 54.1\% & 2.251 & 1.365 \\ \cline{2-6}
  & Both & 74.0\% & 55.4\% & 1.667 & 1.038 \\ \hline
    \multirow{2}{*}{MiniRocket} & Binary & 59.8\% & N/A & 0.667 & 0.395 \\ \cline{2-6}
  & Multi & N/A & 50.4\% & 1.800 & 0.424 \\ \hline

\end{tabular}}
\caption{Validation metrics for model and task combinations, averaged across 5 folds. Train loss is included to measure overfitting}
\label{res}

\end{table}

\subsection{Early Detection Testing}

\begin{figure}[t]
    \centering
    \includegraphics[width=10cm]{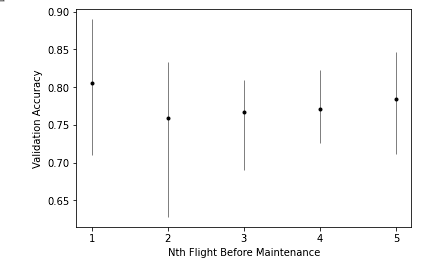}
    \caption{Min, mean, and max early detection accuracy of maintenance issues on validation flights. Only out of fold pre maintenance flights are included in this calculation, done using each fold's best trained model based on validation accuracy. }
    \label{fig:det}
\end{figure}

We can test out of fold early detection performance by creating an early detection dataset, consisting of only validation flights before maintenance in the intake gasket leak/damage class. Note that this does not account for false positives, since they would not be taken in for maintenance. This test is repeated for each fold using the best InceptionTime model, trained on both tasks simultaneously. The results are summarized in Figure \ref{fig:det}. Based on the 5 fold validation, one can conclude that there is no statistically significant difference in recall across the number of flights before maintenance. This suggests that the model is capable of predicting a part failure or the need for maintenance before the problem arises in the flight immediately before maintenance. 

\section{Discussion of Dataset Research Potential}

\paragraph{Flight Safety and Predictive Maintenance Systems}

The Cessna 172 is the most produced aircraft at over 43,000 units \cite{cess} and is widely used in general aviation. This includes flight schools, recreation, and many other applications. However, general aviation is one of the most dangerous forms of civil aviation according to \cite{NTSB}. The NGAFID Aviation Maintenance dataset can be used for training and testing systems that detect maintenance issues early to improve aircraft safety. 

\paragraph{Class Imbalance Research}

One of the major challenges in this dataset is the under-representation of many classes. Half of all flights are recorded after maintenance and two damages classes, intake gasket leak/damage and rocker cover leak/loose/damage, represent the majority of all damage classes. This is so problematic that models trained to predict the specific damage class for a flight will predict that the flight is post maintenance or that the flight belongs to one of the two majority classes more than 95\% of the time. Future researchers can use this dataset to test methods that mitigate problems caused by imbalanced training data. This is particularly relevant for prognostic health management, where it is often easy to gather data of a system in normal operation, but extremely difficult to gather abnormal operation data.  

\paragraph{Contrastive Representation Learning}

Contrastive representation learning is a technique that attempts to learn a representation of the input data in order to provide a benefit to a downstream task, as described by \cite{pmlr-v119-wang20k}. While some flights were excluded from the subset of data used for benchmarking, those flights could be used to learn representations of flight data, which may be useful for improving performance on the downstream detection and classification task. 

\paragraph{Time Series Classification Benchmarking}

Benchmark results suggest that the NGAFID Aviation Maintenance dataset is particularly challenging for non deep learning methods. Of the two main tasks, deep learning models seem to perform reasonable well for maintenance issue detection, but all models perform quite poorly on maintenance issue classification. One possible explanation is that the data is more similar to audio data, where deep methods perform well, and less similar to the time series data benchmarks by \cite{bagnall2018uea}, where methods such as Mini Rocket and HIVECOTE perform well. While we were unable to test HIVECOTE, we invite other researchers to evaluate their methods on the NGAFID Aviation Maintenance dataset. 

\paragraph{Anomaly Detection}

While this dataset does not contain localized annotations of failures, one could separate flights into regular operation and compromised operation and attempt anomaly detection. As noted in earlier sections, the authors of this paper believe that anomaly detection in this dataset would be extremely difficult. This is because the data collected contains a significant degree of variance caused by other factors, such as the pilot's experience, the weather, the flight plan, the payload carried, and so on. 

Furthermore, anomaly detection works best in cases of acute part failure, where the failure of a part is dramatic or has a dramatic impact on the overall system. As noted in earlier sections, chronic part failure is more common and does not create a dramatic impact on the overall system. While such failures make the aircraft unsafe to fly, it does not immediately cause the aircraft to crash or cease operation. 

An auto encoder system, such as \cite{an2015variational}, would struggle with this dataset. First, the encoder must accommodate significant degrees of variance when there are no anomalies. Second, the encoder must detect the most subtle of changes in flight characteristics, which may be easily overshadowed by other sources of variance. Imagine a pilot performing a barrel roll with a leaky gasket; the variance in data caused by the barrel roll would eclipse any variance caused by a leaky gasket. 

\paragraph{Transfer Learning}

This dataset also presents opportunities for transfer learning, \cite{torrey2010transfer}. While this dataset uses sensor data from the Cessna 172 collected from a flight school, the same sensors can be placed on other Cessna 172 aircraft for other uses. For example, the Cessna 172 can be used for passenger, cargo, or military purposes. It is also possible that these sensors can be mounted in similar single engine aircraft. This would allow researchers to develop PHM systems without needing to collect as much data from the other aircraft; the data from the other aircraft can be supplemented with the NGAFID data. 

The authors believe that any transfer learning for a military purpose will not directly endanger more lives. This is because this dataset only contains flight and maintenance information, which can be used to improve the safety and reliability of systems. 
\section{Future Research}

\paragraph{Flight Event Detection Datasets}

Currently, the NGAFID web application provides detection services for certain flight events, such as a aircraft stall event. Using the existing flight data and expert labeling of said flight data, it is possible to create a time series localization dataset, with the goal of detecting both the presence and timing of such events. 

\paragraph{Unsupervised and Self Supervised Datasets}

The NGAFID database contains more than 900,000 hours of flight data from various fleets and aircraft models. While most of the data is unlabeled, it can still be used for self supervised and unsupervised learning, such as contrastive representation learning or masked data modeling, similar to \cite{devlin2018bert}. 

\paragraph{Dataset Expansion}

The authors are actively working with the Federal Aviation Administration and existing flights schools to obtain more maintenance data to expand this dataset to multiple airframes and fleets of aircraft. This process is both slow and costly due to data governance and legal issues. Given the value of this dataset even from a single fleet, the authors will provide the current available data for the Cessna 172 and work on a future data release to include additional air frames and maintenance issues.

\section{Conclusion}

In this paper we present 31,177 hours of flight data across 28,935 flights, which occur relative to 2,111 unplanned maintenance events clustered into 36 types of maintenance issues. Each flight records information from 23 different sensors every second on the Cessna 172 aircraft, during normal operation of a flight school. Our paper makes the significant contribution of providing non-simulated, compromised aircraft flight data, collected ethically at no additional danger to the pilots involved. This dataset is made easily accessible at the links in Section 1. 

The large amount of flight data involving a compromised aircraft is particularly valuable to prognostic health management and predictive maintenance. Because aircraft in question, the Cessna 172, is often used in flight schools, recreation, agriculture, and more, this dataset can help create systems that can greatly improve flight safety. 

Finally, the NGAFID Aviation Maintenance dataset will be of particular interest to machine learning researchers working with time series data. It is our aim, by releasing this dataset and identifying areas for future research, to encourage further work in the detection and classification of maintenance issues. We hope this in turn leads to improved future detection and classification algorithms.

\bibliography{neurips_data_2022}
\bibliographystyle{apalike}


\appendix

\section{Appendix}

\begin{table}[H]
\centering
\small
{\setlength{\extrarowheight}{2pt}%
\begin{tabular}{|l|l|l|}
\hline

  Class Name & Overall Count of Flights & Subset Count of Flights  \\ \hline

aircraft start/external issue                    &          90 &  0 \\ \hline
baffle bracket loose/damage                      &          35 &  0 \\ \hline
baffle crack/damage/loose/miss                   &         545 &  304 \\ \hline
baffle mount loose/damage                        &          42 &  0 \\ \hline
baffle plug need repair/replace                  &         465 &  254 \\ \hline
baffle rivet loose/miss/damage                   &          92 &  0 \\ \hline
baffle screw miss/loose                          &         348 &  211 \\ \hline
baffle seal loose/damage                         &         336 &  197 \\ \hline
baffle spring damage                             &          79 &  0 \\ \hline
baffle tie/tie rod loose or damage               &         303 &  0 \\ \hline
cowling miss/loose/damage                        &          89 &  0 \\ \hline
cylinder compression issue                       &         302 &  143 \\ \hline
cylinder crack/fail/need part repair             &         196 &  108 \\ \hline
cylinder exhaust valve/stuck valve issue         &          48 &  0 \\ \hline
cylinder head/exhaust gas temperature issue      &          71 &  0 \\ \hline
cylinder/exhaust push rod/tube damage            &         106 &  0 \\ \hline
drain line/tube damage                           &         127 &  0 \\ \hline
engine crankcase/crankshaft/firewall near repair &          99 &  0 \\ \hline
engine failure/fire/time out                     &         236 &  161 \\ \hline
engine idle/rpm issue                            &         150 &  93 \\ \hline
engine need repair/reinstall/clean               &         148 &  80 \\ \hline
engine run rough                                 &         311 &  141 \\ \hline
engine seal/tube/bolt loose or damage            &         144 &  76 \\ \hline
engine/propeller overspeed or damage             &         137 &  94 \\ \hline
induction damage/hardware fail                   &          51 &  0 \\ \hline
intake gasket leak/damage                        &        4244 &  2098 \\ \hline
intake tube/bolt/seal/boot loose or damage       &         556 &  269 \\ \hline
magneto failure                                  &          51 &  0 \\ \hline
mixture fail/need adjust                         &          17 &  0 \\ \hline
oil cooler need maintenance                      &         123 &  75 \\ \hline
oil dipstick/tube need repair                    &          20 &  0 \\ \hline
oil leak/pressure issue                          &          53 &  0 \\ \hline
oil return line issue                            &          37 &  0 \\ \hline
pilot/in-flight noticed issue                    &         317 &  75 \\ \hline
rocker cover leak/loose/damage                   &        2157 &  1024 \\ \hline
spark plug need repair/replace                   &          15 &  0 \\ \hline
flights recorded after maintenance & 10291 & 5844 \\ \hline
flights recorded during maintenance & 6504 & 0 \\ \hline

\end{tabular}}

\caption{Count of flights by class in the full dataset and the subset used for the benchmark experiments.}
\label{classes}

\end{table}

\section*{Checklist}


\begin{enumerate}

\item For all authors...
\begin{enumerate}
  \item Do the main claims made in the abstract and introduction accurately reflect the paper's contributions and scope?
    \answerYes{}
  \item Did you describe the limitations of your work?
    \answerYes{Seciton 3}
  \item Did you discuss any potential negative societal impacts of your work?
    \answerNo{The data is collected from purpose built sensors on aircraft. We believe that this data is so specialized as to be only applicable in the aviation industry and only to single engine aircraft, which very few people interact with regularly.}
  \item Have you read the ethics review guidelines and ensured that your paper conforms to them?
    \answerYes{}
\end{enumerate}

\item If you are including theoretical results...
\begin{enumerate}
  \item Did you state the full set of assumptions of all theoretical results?
    \answerNA{}
        \item Did you include complete proofs of all theoretical results?
    \answerNA{}
\end{enumerate}

\item If you ran experiments...
\begin{enumerate}
  \item Did you include the code, data, and instructions needed to reproduce the main experimental results (either in the supplemental material or as a URL)?
    \answerYes{As a repository with links to Colab Notebooks}
  \item Did you specify all the training details (e.g., data splits, hyperparameters, how they were chosen)?
    \answerYes{See the code in the repo}
        \item Did you report error bars (e.g., with respect to the random seed after running experiments multiple times)?
    \answerYes{But only for the early detection in section 4.4. The overall results in 4.3 only report mean values. However, you can reproduce results as the splits are availalbe in the dataset.}
        \item Did you include the total amount of compute and the type of resources used (e.g., type of GPUs, internal cluster, or cloud provider)?
    \answerYes{See section 4.2.5}
\end{enumerate}

\item If you are using existing assets (e.g., code, data, models) or curating/releasing new assets...
\begin{enumerate}
  \item If your work uses existing assets, did you cite the creators?
    \answerYes{}
  \item Did you mention the license of the assets?
    \answerYes{}
  \item Did you include any new assets either in the supplemental material or as a URL?
    \answerYes{}
  \item Did you discuss whether and how consent was obtained from people whose data you're using/curating?
    \answerNo{See data sheet supplementary material}
  \item Did you discuss whether the data you are using/curating contains personally identifiable information or offensive content?
    \answerYes{Section 3.3}
\end{enumerate}

\item If you used crowdsourcing or conducted research with human subjects...
\begin{enumerate}
  \item Did you include the full text of instructions given to participants and screenshots, if applicable?
    \answerNA{}
  \item Did you describe any potential participant risks, with links to Institutional Review Board (IRB) approvals, if applicable?
    \answerNA{}
  \item Did you include the estimated hourly wage paid to participants and the total amount spent on participant compensation?
    \answerNA{}
\end{enumerate}

\end{enumerate}

\end{document}